\documentclass[11pt]{article}

\usepackage[]{acl}

\usepackage{times}
\usepackage{latexsym}
\usepackage{amsmath} 
\usepackage{amssymb}
\usepackage{booktabs} 
\usepackage{multirow}
\usepackage{makecell}
\usepackage{cuted}
\usepackage{caption}
\usepackage{tabularx}

\usepackage{graphicx}
\usepackage{subcaption}  
\usepackage[font=small, labelfont=bf]{caption} 

\usepackage[T1]{fontenc}

\usepackage[utf8]{inputenc}

\usepackage{microtype}

\usepackage{inconsolata}

\usepackage{graphicx}

\title{Reinforcement Learning for Compositional Generalization with Outcome-Level Optimization}

\author{Xiyan Fu$^{1}$  and Wei Liu$^{2}$ \\
$^{1}$ Nanyang Technological University $^{2}$ Independent Researcher \\
\texttt{xiyan.fu@ntu.edu.sg | wei.liu.llm@gmail.com}
}

\begin{document}
\maketitle
\begin{abstract}
Compositional generalization refers to correctly interpret novel combinations of known primitives, which remains a major challenge. Existing approaches often rely on supervised fine-tuning, which encourages models to imitate target outputs. This token-level training paradigm fails to capture the global compositional structure required for generalizing to unseen combinations. In this work, we investigate whether compositional generalization can instead be improved through outcome-level reinforcement learning. We adopt Group Relative Policy Optimization to optimize models based on feedback on their final outputs. Within this framework, we explore both a simple binary outcome reward and a composite reward that provides additional composition feedback. Experiments on multiple compositional benchmarks show that reinforcement learning improves compositional generalization compared to supervised fine-tuning. Further analysis reveals that supervised models tend to overfit frequent training compositions, whereas reinforcement learning improves compositional generalization by reshaping the output distribution, particularly for more complex composition types.
\end{abstract}

\section{Introduction}
\begin{figure}
    \centering
    \includegraphics[width=1\linewidth]{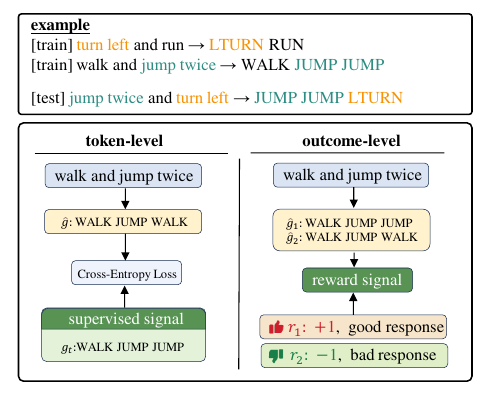}
    \caption{Illustration of compositional generalization and training paradigms. \textit{Top}: Example of compositional generalization where the model must correctly compose previously seen primitives (e.g., jump twice and turn left) to produce the correct action sequence for a novel instruction. \textit{Bottom}: Comparison of training signals. Token-level optimization relies on supervised targets and cross-entropy loss, whereas outcome-level optimization samples multiple responses and optimizes them using reward signals.}
    \label{fig:intro}
\end{figure}

Compositional generalization \citep{fodor1988connectionism, lake2017building, hupkes2020compositionality}, a hallmark of human cognition, refers to the ability to understand and produce novel combinations by systematically recombining known components. As illustrated in Figure~\ref{fig:intro}, humans who understand `turn left and run $\rightarrow$ LTURN RUN' and `walk and jump twice $\rightarrow$ WALK JUMP JUMP' can infer a new combination `jump twice and turn left $\rightarrow$ JUMP JUMP LTURN' by recombining the known components. This ability enables extrapolation beyond the empirical data distribution and is therefore widely regarded as essential for models to generalize to unseen and increasingly complex reasoning tasks.

Given the importance of compositional generalization, extensive research has explored how to equip neural models with this capability. Early works \citep{lake2018generalization,kim-linzen-2020-cogs} mainly relied on supervised fine-tuning (SFT) with teacher forcing~\citep{lamb2016professor}, where models are trained to imitate target outputs. However, such methods often struggle to generalize to novel compositions despite strong in-distribution performance. Subsequent studies have attempted to address this limitation through data augmentation \citep{andreas-2020-good,li-etal-2023-learning,yao-koller-2024-simple}, which exposes models to diverse compositional patterns, or architecture designs \citep{guo2020hierarchical} that introduce compositional inductive biases. While these approaches improve performance on compositional benchmarks, they still rely on token-level supervision that optimizes local token predictions. However, compositional correctness is inherently a global property of the output sequence, i.e., an output may appear locally plausible while still violating the intended composition rule. This mismatch raises a natural question: \textit{can compositional generalization be learned through outcome-level optimization} instead?





To explore this question, we adopt a reinforcement learning (RL) framework, in which models are optimized based on feedback on their final outputs rather than token-level supervision. 
Specifically, we employ Group Relative Policy Optimization (GRPO)~\citep{deepseek-math} as the training framework. Figure~\ref{fig:intro} compares the training signals used in SFT and GRPO. Within this framework, we design two types of reward signals: (i) a coarse-grained binary reward that provides feedback based on the exact match between the generated output and the ground truth, and (ii) a fine-grained composite reward that decomposes the task and provides additional feedback for primitive prediction and compositional structure. Across multiple compositional benchmarks, we find that RL training improves compositional generalization performance. Interestingly, we observe that two rewards achieve comparable results, suggesting that compositional generalization may emerge from minimal outcome-level feedback.

To better understand differences between token-level supervision and outcome-level optimization in composition, we analyze the behaviors of models trained with SFT and RL. We investigate performance across compositional types, finding larger improvements from RL on productivity tests (i.e., length splits), particularly for compositions closer to the training length. Furthermore, our analysis reveals that SFT-trained models tend to overfit compositional patterns observed during training, often reproducing familiar structures when they make incorrect predictions for novel compositions. In contrast, RL-trained models reshape the output probability distribution, reducing the bias toward frequent training compositions and improving the ranking of candidates. This behavior becomes evident under multi-sample evaluation (e.g., pass@k), where RL mainly improves top-1 accuracy while achieving comparable performance at larger $k$.

Our main contributions are as following:
\begin{itemize}
    \item We propose reinforcement learning paradigm for compositional generalization, where models are optimized using feedback on the final output rather than token-level supervision.
    \item We design two types of reward signals, binary and composite, to guide RL training and improve compositional generalization across multiple benchmarks.
    \item Through behavioral analyses, we demonstrate that SFT tends to overfit training patterns. In contrast, RL reshapes the output distribution and leads to improvements in compositional generalization ability, especially on complex composition types.
\end{itemize}

\section{Related Work}
\paragraph{Compositional Generalization} 
Compositional generalization refers to the ability of a model to systematically recombine known primitives or rules to understand or generate novel compositions \citep{fodor1988connectionism, lake2017building, hupkes2020compositionality}. It has been extensively studied across diverse settings, such as in semantic parsing \citep{lake2018generalization,kim-linzen-2020-cogs,qiu-etal-2022-evaluating}, machine translation \citep{li-etal-2021-compositional, dankers-etal-2022-paradox}, deductive reasoning \citep{saparov2023testing}, natural language inference \citep{yanaka-etal-2020-neural, fu-frank-2023-seti}, etc. They found that even state-of-the-art LLMs are still not able to perform compositional generalization. 

To address this limitation, a number of approaches have been proposed. \citet{qiu-etal-2022-improving, levy-etal-2023-diverse, yao-koller-2024-simple} introduce data augmentation strategies that inject compositional inductive bias into neural sequence models by generating additional training examples or recombining components across existing examples. Other works \citep{zheng-lapata-2021-compositional-generalization, herzig-berant-2021-span} design specialized model architectures to better capture compositional structures, such as modular architectures or parallel decoding mechanisms. More recent studies explore alternative training paradigms, including meta-learning \citep{conklin-etal-2021-meta, lake2023human} and continual learning \citep{fu-frank-2024-exploring}, to improve models' adaptability to more realistic scenarios. However, these approaches still rely on token-level supervision, optimizing models to predict individual tokens rather than ensuring the correctness of the entire compositional output sequence. In contrast, we investigate whether compositional generalization can be improved through reinforcement learning, which provides outcome-level feedback and enables global optimization over complete sequences.


\paragraph{Reinforcement Learning for Generalization}

Reinforcement learning is widely used in the post-training of LLMs to better align model outputs with human preferences \citep{ouyang2022, achiam2023gpt}. Beyond alignment, recent studies suggest that RL can also improve generalization compared to imitative learning \citep{bi2024deepseek, lambert2025tulu}. In particular, RL tuning enhances cross-lingual generalization, especially when trained on non-English data \citep{huang2026beyond}, as well as cross-category generalization, where training on one domain (e.g., geometry) transfers to others (e.g., algebra) \citep{wang2025reinforcement}. It also improves cross-domain generalization, e.g., \citet{wei2025swerl} introduce Llama3-SWE-RL, which achieves strong performance on software engineering tasks despite being trained solely on software evolution data. Similar trends are observed in multimodal settings, where \citet{chu2026gpg} report improved generalization across both unimodal and multimodal reasoning benchmarks. More recently, several studies have explored modifications to the RL training paradigm to further enhance generalization. For example, \citet{tang2025rl,zhang2026onpolicy} combine on-policy and off-policy learning to improve the generalizability of reasoning language models. 

In contrast to these works, we focus on compositional generalization, investigating how RL affects the ability of models to handle novel combinations of known primitives. In addition, we conduct a detailed comparison between RL and supervised fine-tuning (SFT), showing that RL reshapes distributions and avoids copying frequent patterns of training compositions.

\section{Methods}
\subsection{Problem Definition}
Compositional generalization tests are designed to evaluate whether a model can generalize to unseen compositions whose constituting primitives and composing rules have been observed in training. For example, we can evaluate a model’s compositional generalization ability by testing it on an unseen compositional sample \textit{jump twice and turn left $\rightarrow$ JUMP JUMP LTURN} after training with \textit{jump twice after run $\rightarrow$ RUN JUMP JUMP} and \textit{run and turn left $\rightarrow$ RUN LTURN}. Here, \textit{jump $\rightarrow$ JUMP} and \textit{turn left $\rightarrow$ LTURN} are primitives, \textit{$v_x$ and $v_y$ $\rightarrow$ X Y} is the implicit composition rule. We denote the set of primitives as $\mathcal{P}$ and the set of composition rules as $\mathcal{R}$. A compositional instance is generated by applying a composition rule $r \in \mathcal{R}$ to a tuple of primitives $(p_1, \dots, p_k)$, where $p_i \in \mathcal{P}$. Each compositional type can therefore be represented as a pair $(\mathbf{p}, r)$, where $\mathbf{p}$ denotes the primitives involved. Correspondingly, $\mathcal{C}$ denotes the set of all possible compositional types:
\begin{equation}
    \mathcal{C} = \{(\mathbf{p}, r) \mid \mathbf{p} \subseteq \mathcal{P},\ r \in \mathcal{R}\}.
\end{equation}

\noindent For compositional generalization tests, the sets of compositional types used for training and testing are always ensured disjoint, $\mathcal{C}_{\text{train}} \cap \mathcal{C}_{\text{test}} = \emptyset$. At the same time, all primitives and composition rules appearing in test instances are guaranteed to have been observed during training: $\mathcal{P}_{\text{test}} \subseteq \mathcal{P}_{\text{train}}$, $\mathcal{R}_{\text{test}} \subseteq \mathcal{R}_{\text{train}}$. This setup ensures that test instances require genuine recombination of known primitives under familiar composition rules, rather than extrapolation to entirely unseen components.

\subsection{Compositional Group Relative Policy Optimization}

\begin{figure*}
    \centering
    \includegraphics[width=\linewidth]{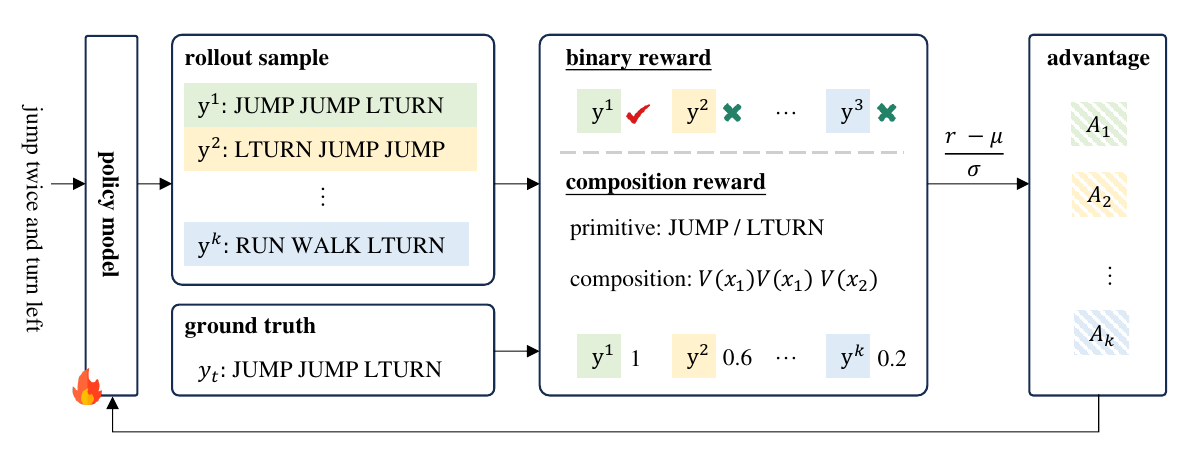}
    \caption{Overview of Compositional Group Relative Policy Optimization. The LLM samples a group of candidate label sequences $(y^1, y^2, \ldots, y^K)$ based on inputs. Each candidate is then evaluated using two complementary reward signals: (i) a \textit{binary reward} that measures exact match with the gold sequence, and (ii) a \textit{compositional reward} that assesses primitive correctness and structural composition patterns. Rewards are normalized within the sampled group to compute relative advantages, which are then used to update the policy via group-relative optimization. }
    \label{fig:model}
\end{figure*}

To investigate whether compositional generalization can be learned through outcome-level optimization, we adopt a reinforcement learning (RL) training framework. This paradigm allows LLMs to autonomously explore and discover latent compositional structures through reward-based feedback. Specifically, we employ Group Relative Policy Optimization (GRPO) \citep{deepseek-math}, a computationally efficient RL algorithm that has demonstrated strong empirical performance. The overall training procedure is illustrated in Figure~\ref{fig:model}.

\subsubsection{Policy and Rollout Generation}
We parameterize the language model as a stochastic policy $\pi_\theta(y \mid x)$, which generates an output sequence $y$ conditioned on an input $x$. For each training input, we sample a group of $K$ candidate outputs, forming a rollout group $\mathcal{G}$ associated with the same input.
\begin{equation}
\mathcal{G} = \{y^{1}, \dots, y^{K}\} \sim \pi_\theta(\cdot \mid x),
\end{equation}
Each output $y^i$ in the group is a complete compositional prediction generated by the model.

\subsubsection{Reward Assignment}
For each sampled output $y^{i}$, we assign a scalar reward $r^{i}$ based on the correctness of the generated prediction. We consider two types of reward functions: a \textit{binary reward} that provides coarse-grained supervision, and a \textit{composite reward} that offers more fine-grained feedback.

\paragraph{Binary Reward.}
The binary reward assigns a score of either $0$ or $1$ based on exact matching between the generated compositional prediction and the ground-truth output. While simple and deterministic, this binary reward treats all incorrect trajectories uniformly, failing to capture partial correctness or relative quality among different erroneous predictions. \par

\paragraph{Composite Reward.}
To provide more fine-grained feedback for incorrect predictions, we introduce a composite reward that evaluates both \emph{primitive correctness} and \emph{compositional structure}. Given a predicted action sequence $y^{pred}$ and a ground-truth sequence $y^{gold}$, we define the two reward components as follows:

\begin{itemize}
    \item \textbf{Primitive.}
    We extract the sets of primitives appearing in the predicted and gold sequences, denoted as $P^{\text{pred}}$ and $P^{\text{gold}}$, respectively. The primitive reward measures the coverage of gold primitives by the prediction, denoted as:
    \begin{equation}
    R_{\text{prim}}(y^{\text{pred}}, y^{\text{gold}})
    =
    \frac{\left| P^{\text{pred}} \cap P^{\text{gold}} \right|}
         {\left| P^{\text{gold}} \right|}.
    \label{eq:prim}
    \end{equation}
    This reward encourages the model to correctly identify and generate the required primitives, while ignoring invariant to their ordering and repetition. 
    For example, in the compositional instruction example, \textit{jump twice and turn left $\rightarrow$ JUMP JUMP LTURN}, the primitives \textsc{JUMP} and \textsc{LTURN} are extracted.

    \item \textbf{Composition.}
    We abstract away concrete primitive identities and focus solely on their compositional patterns. Each primitive is mapped to a variable, yielding a normalized and ordered sequence that represents the compositional skeleton. We then compute the compositional reward by measuring position-wise matches between the predicted and gold skeletons. Let $C^{\text{pred}}$ and $C^{\text{gold}}$ denote the skeletons derived from the predicted and gold sequences, respectively. The compositional reward is defined as:
    \begin{equation}
    R_{\text{comp}}(y^{\text{pred}}, y^{\text{gold}})
    =
    \frac{1}{\left|C^{\text{gold}}\right|}
    \sum_{i=1}^{\left|C^{\text{gold}}\right|}
    \mathbf{1}\!\left[
    C^{\text{pred}}_i = C^{\text{gold}}_i
    \right]
    \label{eq:compos}
    \end{equation}
    This reward encourages the model to focus on compositional structures, while ignoring primitives. For instance, the instruction \textit{jump twice and turn left $\rightarrow$ JUMP JUMP LTURN} yields the compositional skeleton $V(x_{1})$ $V(x_{1})$ $V(x_{2})$, obtained by mapping each distinct primitive to a unique placeholder variable.
\end{itemize}

The final composite reward \footnote{The specific definitions of primitives and compositions for each dataset are provided in Table~\ref{tab:datasets} (Appendix~\ref{app:compos_example}).} is obtained as the weighted sum of the primitive and composition components. We denoted it as $R = \lambda_1 R_{\text{prim}} + \lambda_2 R_{\text{comp}}$, where $\lambda_1$ and $\lambda_2$ are hyper-parameters.

\subsection{Group Relative Policy Optimization}
To optimize the policy, we employ Group Relative Policy Optimization (GRPO), where the advantage is computed by comparing the relative performance of trajectories within the same sampled group $\mathcal{G}$. Specifically, we compute a normalized advantage for each sampled output by subtracting the group mean reward:
\begin{equation}
\hat{A}_i = \frac{r_i - \mu_r}{\sigma_r}.
\label{eq:adv}
\end{equation}
where $\mu_r$ and $\sigma_r$ denote the mean and standard deviation of rewards within the rollout group. This group-relative advantage emphasizes which generations perform better or worse compared to other candidates under the same input, mitigating reward scale sensitivity and stabilizing training.

Given the computed advantages, the policy is updated by maximizing the expected advantage-weighted log-likelihood of generated outputs while constraining excessive deviation from a reference policy. The GRPO objective is defined as:
\begin{equation}
\begin{aligned}
\mathcal{J}_{\text{GRPO}}(\theta)
= & \ \mathbb{E} \Bigg[
\frac{1}{K} \sum_{i=1}^{K}
\frac{1}{|y^i|}
\sum_{t}
\mathcal{L}_{\text{clip}}
\big(r_{i,t}(\theta), \hat{A}_{i,t}\big)
\Bigg] \\
& - \beta D_{\mathrm{KL}}(\pi_\theta \| \pi_{\mathrm{ref}})
\end{aligned}
\end{equation}

\noindent The clipping operator limits excessively large policy updates with threshold $\epsilon$, denoted as:
\begin{equation}
\mathcal{L}_{\text{clip}}(r,\hat{A})
=
\min(r\hat{A}, \text{clip}(r,1-\epsilon,1+\epsilon)\hat{A})
\end{equation}

\noindent Here, $K$ denotes the number of sampled outputs in each group, and $|y^i|$ represents the length of $i$-th generated output $y^i$. The term $r_{i,t}(\theta)$ denotes the probability ratio between the current policy $\pi_\theta$ and the behavior policy used to generate the samples, while $\hat{A}_{i,t}$ represents the token-level advantage defined in Eq. \ref{eq:adv}. The KL divergence with strength $\beta$ regularizes the policy, preventing excessive deviation from the supervised distribution.

\begin{table*}
    \centering
    \resizebox{\textwidth}{!}{
    \begin{tabular}{l|cc|c|cc|cccc} \toprule
          \multirow{2}{*}{} &\multicolumn{2}{c|}{SCAN} & \multirow{2}{*}{COGS} & \multicolumn{2}{c|}{GeoQuery} & \multicolumn{4}{c}{CFQ}   \\ \cmidrule(r){2-3}  \cmidrule(r){5-6} \cmidrule(r){7-10}
        &length &  turn left & & template &length & MCD1 & MCD2 & MCD3 & Avg \\ \midrule
        \textit{Llama} \\
         SFT &11.88 &69.03 &81.12 &50.36 & 43.21 &91.04 &74.85 &77.42 &81.10  \\
         GRPO-Binary  & 18.49 & 71.52 &83.90  &52.48 &45.92 &91.35 &76.44 &81.46& 83.08\\
         GRPO-Compos   &23.44 &70.74 &82.88 &52.07 &46.17 &92.10 &76.16 &83.44& 83.90\\ \midrule
         \textit{Qwen} \\
         SFT  &18.41 &78.06 &82.09 &73.65 &28.21 &88.52  &73.98 &73.67 &78.72 \\
         GRPO-Binary  &22.68 &79.13  &83.23 &75.27 &32.14 &89.23 &75.73 &76.54 &80.50\\
         GRPO-Compos   &21.81 &78.76 &83.15 &77.90 &33.76 &89.16 &77.33 &75.86 &80.78\\\bottomrule
    \end{tabular}}
    \caption{Results of SFT and GRPO on four compositional generalization benchmarks across various settings. We report exact matching accuracy for all evaluations. All results are averaged over three runs and reported as mean accuracy.}
    \label{tab:main_rst}
\end{table*}

\section{Setup}
\subsection{Datasets}
\label{ref:sec_data}
We evaluate on four standard compositional generalization benchmarks: SCAN, COGS, GeoQuery, and CFQ. They capture complementary aspects of compositionality and form a diverse testbed for evaluation. Data statistics and examples of primitive and compositional structures for reward design are provided in Appendix \ref{app:statistics} and \ref{app:compos_example}. 

\paragraph{SCAN} \citep{lake2018generalization} is a synthetic dataset designed to test compositional generalization in grounded language understanding. Each example pairs a simple navigation command with a sequence of atomic actions representing its execution, e.g., \textit{jump twice and turn left $\rightarrow$ JUMP JUMP LTURN}. For composite reward, we define primitives at the action level (e.g., JUMP, LTURN), while compositional structures are abstracted as higher-level templates over primitives (e.g., $V(x_{1})$ $V(x_{1})$ $V(x_{2})$). SCAN provides several splits targeting different compositional generalization settings. We select the \textit{turnleft} and \textit{length} splits, as they are more challenging \citep{qiu-etal-2022-improving}.

\paragraph{COGS} \citep{kim-linzen-2020-cogs} is a benchmark designed to evaluate compositional generalization in semantic interpretation. The task is to map natural language sentences to structured semantic representations, e.g., \textit{A hedgehog ate the cake $\rightarrow$ \*cake($x_4$) ; hedgehog($x_1$) AND eat.agent($x_2$, $x_1$) AND eat.theme($x_2$, $x_4$)}. For composite reward, we define predicates and entities as primitives (e.g., hedgehog, cake, eat.agent, eat.theme) and abstract their underlying argument structures into compositional templates (e.g., $N(x_4) \;\wedge\; N(x_1) \;\wedge\; V.agent(x_2, x_1) \;\wedge\; V.theme(x_2, x_4) $). 

\paragraph{GeoQuery} \citep{herzig-berant-2021-span} is a semantic parsing benchmark that maps natural language questions to logical forms representing queries over a geographic database. Each example consists of a question paired with a corresponding structured query, e.g., \textit{What states border Texas? $\rightarrow$ answer state next\_to stateid texas}. For compositional reward design, we decompose logical forms into atomic primitives, including entities and relations (e.g., stateid(texas), next\_to), and abstract their compositional templates (e.g., $N(x_1) \;\wedge\; R(x_1, x_2)$). We use the linearized FunQL representation provided in the dataset, where logical forms are expressed as sequences of function tokens.

\paragraph{CFQ} \citep{keysers2020measuring} is a semantic parsing benchmark for evaluating compositional generalization in mapping natural language questions to SPARQL queries. Each instance consists of a question paired with its corresponding logical form, e.g., \textit{Who directed Elysium? $\rightarrow$ SELECT DISTINCT ?x0 WHERE \{?x0 a ns:people.person. ?x0 ns:film.director.film m.0gwm\_wy .\}}. For composite reward, we further decompose SPARQL queries into atomic primitives as entities and relations (e.g., ns:people.person, ns:film.director.film), and abstract the composition structures (e.g., $N(x_1) \;\wedge\; R(x_1, x_2)$). We adopt the three Maximum Compound Divergence (MCD) splits provided in the original dataset, which are specifically designed to induce strong compositional distribution shifts.


\subsection{Implementation Details}
We conduct our experiments on Qwen-2.5-7B-Instruct \citep{qwen2.5} and Llama-3.1-8B-Instruct \citep{grattafiori2024llama}, given their superior performance. During training, we observe that reinforcement learning from scratch suffers from unstable optimization due to the large and sparse exploration space. Therefore, we adopt a supervised warm-up stage via SFT to initialize the policy following previous works \citep{deepseek-math}, which significantly stabilizes subsequent GRPO training.\footnote{For GRPO training, we use the code \url{https://github.com/verl-project/verl}.} After the warm-up stage, we further optimize the policy using Compositional Group Relative Policy Optimization. For each input, we sample 8 trajectories with a temperature of 0.6. The update batch size is 8, and the learning rate is set to 1e-6. The hyperparameters $\lambda_1$ and $\lambda_2$ for the composite reward are set to 0.1 and 0.2, respectively.

\section{Results}
\subsection{Main Results}
Table \ref{tab:main_rst} reports the compositional generalization results across four benchmarks using two base models. Compared with the SFT baseline, both GRPO variants, GRPO-Binary and GRPO-Composite, consistently achieve better performance across all benchmarks and model bases. The results suggest that \textit{outcome-level optimization (i.e., RL-based strategies) is more effective than token-level imitation (i.e., SFT-based strategies) for learning compositional behaviors}. 

We further analyze the results across different types of compositional generalization.\footnote{Different compositional generalization types are discussed in \citet{hupkes2020compositionality}} We observe that the improvements brought by GRPO are generally larger on productivity settings that require generating longer compositions of known primitives. For example, on the SCAN length split, GRPO-Binary improves performance by 6.61\% and 4.27\% for the two base models, respectively. In contrast, the gains are smaller on systematicity evaluations, where generalization requires recombining known primitives into novel structures. For instance, on the MCD3 split of CFQ, which exhibits the largest generalization gap, the improvements are 4.04\% and 2.87\%. Similar trends are observed across GRPO-Composite strategies. Notably, the absolute performance on productivity tasks (e.g., length splits) is substantially lower than that on systematicity evaluations, indicating that productivity compositional generalization is considerably more challenging. These results suggest that outcome-level optimization exhibits different benefits across compositional generalization types and is particularly effective for more challenging settings such as productivity tasks.

In addition, we compare different GRPO reward strategies and observe that GRPO-Binary and GRPO-Compos achieve comparable performance across various settings. This suggests that compositional generalization tests are relatively insensitive to the specific reward formulation. That is, a simple binary outcome signal is already sufficient to guide effective policy optimization.

\begin{table}
    \centering
    \resizebox{\columnwidth}{!}{
    \begin{tabular}{lllllll} \toprule
     & \multicolumn{3}{c}{CFQ-MCD3} & \multicolumn{3}{c}{SCAN-Length} \\ \cmidrule(r){2-4} \cmidrule(r){5-7}
     &  prim  & com & em  &  prim  & com & em \\ \midrule
     $R_{P+C}$&98.66& 90.46 & 83.44 & 97.96 & 92.41 & 23.44    \\
     $R_{P}$ &98.58 &88.43 &81.60 &97.81 &89.72 & 20.97  \\
     $R_{C}$ &98.32 &90.19 &82.73 &97.19 &92.15 & 22.13 \\ \bottomrule
    \end{tabular}
    }
    \caption{Ablation study of GRPO-Compos training paradigm based on Llama-3.1-Instruct. We decompose the composite reward design $R_{P+C}$ by only offering primitive reward $R_P$ and composite reward $R_C$. \textit{prim} and \textit{com} represent primitive and composition accuracy, respectively. \textit{em} denotes the exact matching accuracy.}
    \label{tab:ablation}
\end{table}

\subsection{Ablation Study}
In this section, we conduct an ablation study to investigate the effect of compositional reward design in the GRPO-Compos training paradigm on two challenging compositional generalization settings, i.e., MCD3 of CFQ and length of SCAN. Specifically, we decompose the composite reward into its individual components and train GRPO with only one reward at a time: i) the primitive reward, which provides positive feedback for correctly predicting primitives, and ii) the composition reward, which rewards correct compositional structures. The full GRPO-Compos model combines both signals. Table \ref{tab:ablation} reports the ablation results.

We observe that the full composite reward consistently achieves higher exact-match accuracy than using either reward alone. This indicates that the two reward signals provide complementary information for learning compositional generalization. 
To better understand the effect of different reward signals, we further decompose the evaluation metrics into primitive accuracy and composition accuracy. Their definitions are provided in Eq.~\ref{eq:prim} and Eq.~\ref{eq:compos}. As shown in Table \ref{tab:ablation}, primitive reward primarily improves primitive accuracy, while composition reward contributes more directly to composition accuracy. Interestingly, primitive accuracy remains comparable between primitive reward only ($R_P$) and composite reward only ($R_C$), suggesting that composition reward can also indirectly enhance primitive prediction. Overall, these results highlight a non-trivial interaction between reward design and the learning dynamics in RL.

\begin{figure}
    \centering
    \includegraphics[width=1\linewidth]{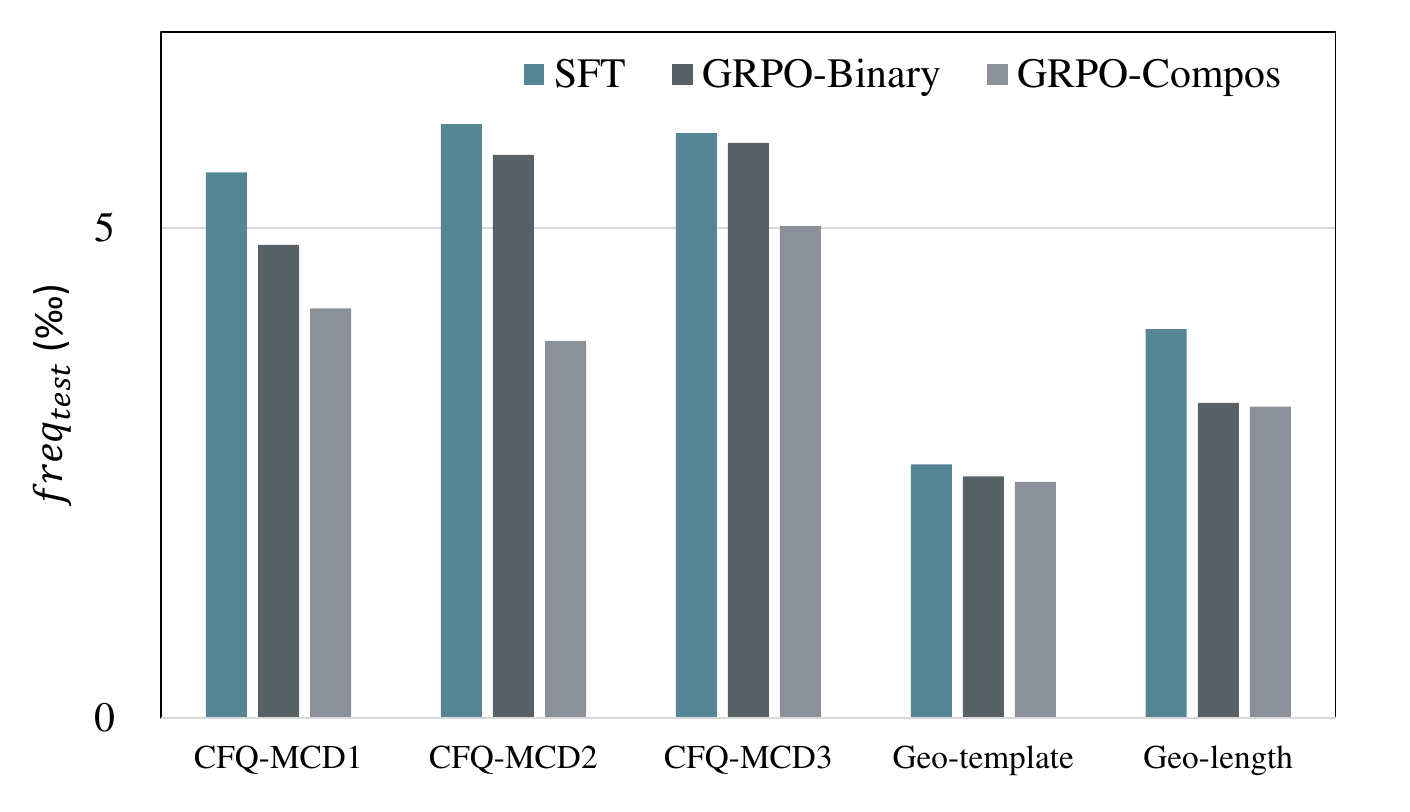}
    \caption{Average training trigram frequency of incorrect predictions from SFT and GRPO. Trigram frequencies are computed with respect to the training data, excluding trigrams appearing in ground-truth outputs.}
    \label{fig:overfit}
\end{figure}

\section{Analysis}
In this section, we further compare results from the SFT and GRPO training paradigms to better understand how they influence compositional generalization. 

\subsection{Copying Behavior}
Prior works \citep{pmlr-v119-rice20a} suggest that poor generalization may stem from models overfitting to the prior output distribution of the training data. In compositional generalization tasks, this manifests as models relying on frequently occurring output fragments rather than composing primitives according to the input. In this section, we examine whether token-level optimization (i.e., SFT) encourages models to follow such frequent structural patterns, and whether outcome-level optimization (i.e., GRPO) can alleviate this issue.

To quantify this tendency, we examine whether model predictions prefer output trigrams that are more frequent in the training data. We first compute trigram statistics from the output sequences in the training set and normalize their frequencies, denoted as $freq_{train}$. For incorrect samples, we extract all trigrams from the model predictions and look up their frequencies according to $freq_{train}$. We then define the average trigram frequency of a sequence as a measure of how strongly it aligns with common structural patterns in the training data. To avoid bias from valid output patterns, we exclude trigrams that also appear in the ground-truth outputs when computing the average trigram frequency. Finally, we compare this metric for incorrect samples produced by SFT and GRPO. We conduct this analysis on CFQ and GeoQuery, where trigrams capture meaningful structural patterns. The results are shown in Figure~\ref{fig:overfit}.

We observe that the SFT paradigm consistently produces predictions with higher average trigram frequency than GRPO. This suggests that SFT tends to rely more on frequent structural patterns from the training data, whereas GRPO exhibits weaker alignment with such patterns. Furthermore, we find that GRPO-Compos yields lower trigram frequencies than GRPO-Binary, particularly on the CFQ benchmark, even though it does not lead to higher accuracy (see Table~\ref{tab:main_rst}). This result suggests that rewarding primitives and compositional structures helps mitigate the tendency to copy frequent patterns from the training data.

\begin{table}
    \centering
    \resizebox{\columnwidth}{!}{
    \begin{tabular}{lllllll} \toprule
     & \multicolumn{3}{c}{CFQ-MCD3} & \multicolumn{3}{c}{SCAN-Length} \\ \cmidrule(r){2-4} \cmidrule(r){5-7}
     &  @1  & @5 & @10  &  @1  & @5 & @10 \\ \midrule
     SFT&77.90 &82.21 &83.28 &11.02& 20.69  &23.77   \\
     GRPO-B &81.57 &82.66 &83.42 &18.57 &21.10 &23.81 \\
     GRPO-C &83.42 &83.53 &83.61 &23.85 &24.21 &24.39 \\ \bottomrule
    \end{tabular}
    }
    \caption{Comparison of top-k accuracy (pass@k) for SFT and GRPO-based methods, i.e., GRPO-B(inary) and GRPO-C(omposite).}
    \label{tab:passk}
\end{table}

\subsection{Top-k Performance}
The trigram analysis suggests that SFT tends to rely more heavily on frequent patterns from the training data. Such reliance may affect the ranking of candidate outputs during decoding, causing inferior compositional generalization ability. To further examine this effect, we evaluate models using top-k accuracy. Specifically, for each test example, we set the sampling temperature to 0.6 and generate $k$ candidate outputs. We then measure whether at least one sampled output matches the gold output, denoted as \textit{pass@k}. Formally, $\text{pass@k} = \frac{1}{N} \sum_{i=1}^{N} \mathbf{1}\!\left[\exists j \le k,\ \hat{y}_i^{(j)} = y_i \right]$, where $y_i$ denotes the gold output and $\hat{y}_i^{(j)}$ is the $j$-th sampled prediction. Table \ref{tab:passk} presents results.

We find that GRPO variants achieve higher top-1 accuracy (pass@1) than SFT on both benchmarks, while their top-10 performance (pass@10) remains comparable. This suggests that GRPO mainly improves the ranking of candidate outputs by assigning higher probability to correct structures, resulting in a sharper output distribution. In contrast, SFT tends to rank frequent training patterns higher, which may lead to incorrect predictions being prioritized during decoding. Consistently, the larger gap between pass@1 and pass@k of SFT indicates greater sensitivity to the choice of $k$, whereas GRPO remains comparatively stable across different decoding budgets.

\begin{figure}
    \centering
    \includegraphics[width=1\linewidth]{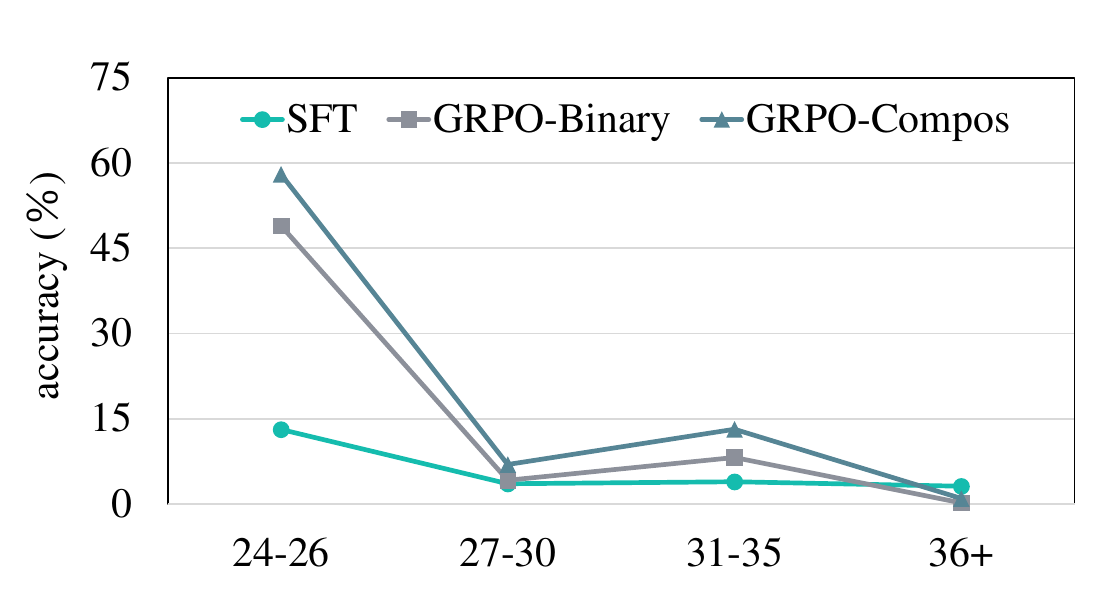}
    \caption{Performance on the SCAN-Length split across various target output lengths. Examples are grouped into bins by output length.}
    \label{fig:length}
\end{figure}

\subsection{Length Extrapolation}
To further understand how different training paradigms behave under varying compositional complexity, we analyze model performance with respect to output length. In compositional generalization tasks, longer outputs typically correspond to more complex structural compositions and therefore present a more challenging compositionality. We group the test examples into several buckets according to the gold output length (24–26, 27–30, 31–35, and 36+) and report the corresponding pass@k performance for each bucket.


Figure~\ref{fig:length} shows the accuracy of SFT and GRPO on SCAN-length as output length increases. All methods exhibit a general decline beyond lengths of 24–26, indicating that compositional generalization degrades with longer outputs and remains challenging even under outcome-level optimization. GRPO consistently outperforms SFT, particularly on shorter outputs (24–26 tokens), suggesting that SFT struggles to produce structurally correct outputs even at shorter lengths. In contrast, GRPO mitigates this issue through reward-based optimization, encouraging structurally valid predictions. Moreover, GRPO-Compos achieves slightly better performance, indicating that explicitly rewarding primitives and compositional structures further enhances the model’s ability to capture compositional patterns.

\section{Conclusion}
In this work, we study whether compositional generalization can be improved through outcome-level optimization rather than token-level supervised training. Experiments across multiple benchmarks show that GRPO consistently improves compositional generalization. Additionally, we find that a simple binary reward performs comparably to composite rewards. Further analysis suggests that RL reshapes the distribution and mitigates overfitting to training compositions, thereby improving compositional generalization.

\section{Limitation}
Reinforcement learning introduces additional computational cost compared to standard supervised fine-tuning, which may limit its practicality in large-scale training scenarios. In addition, our study only explores relatively simple reward formulations, including binary and composite rewards. More sophisticated reward designs or alternative outcome-level feedback signals may further affect the effectiveness of reinforcement learning for compositional generalization, which we leave for future work.

\bibliography{custom,anthology}

\clearpage

\appendix
\begin{strip}
    \centering
    \resizebox{\textwidth}{!}{
    \begin{tabular}{lll}
        \toprule
        Dataset & Setting & Example \\
        \midrule
        \multirow{3}{*}{SCAN} &original  & jump twice and turn left $\rightarrow$ JUMP JUMP LTURN\\  \cmidrule(r){2-3}
         &primitive  & JUMP LTURN\\ \cmidrule(r){2-3}
         &composition  & $V(x_{1})$ $V(x_{1})$ $V(x_{2})$ \\ \midrule
         \multirow{3}{*}{COGS} &original  & \makecell[l]{A hedgehog ate the cake $\rightarrow$ \\ \*cake($x_4$) ; hedgehog($x_1$) AND eat.agent($x_2$, $x_1$) AND eat.theme($x_2$, $x_4$) }\\ \cmidrule(r){2-3}
         &primitive  & \makecell[l]{[\textit{entity prim}] hedgehog, cake  [\textit{relation prim}] eat.agent, eat.theme} \\ \cmidrule(r){2-3}
         &composition  & $N(x_4) \;\wedge\; N(x_1) \;\wedge\; V.agent(x_2, x_1) \;\wedge\; V.theme(x_2, x_4) $\\ \midrule
         \multirow{3}{*}{GeoQuery} &original  & \makecell[l]{What states border Texas? $\rightarrow$ answer state next\_to stateid texas }\\ \cmidrule(r){2-3}
         &primitive  & \makecell[l]{[\textit{entity prim}] texas [\textit{relation prim}] next\_to} \\ \cmidrule(r){2-3}
         &composition  &\makecell[l]{$N(x_1) \;\wedge\; R(x_2, x_1)$ } \\ \midrule
        \multirow{3}{*}{CFQ} &original  & \makecell[l]{Who directed Elysium? $\rightarrow$ \\ SELECT DISTINCT ?x0 WHERE \{ \\ \quad ?x0 a ns:people.person. \\ \quad ?x0 ns:film.director.film m.0gwm\_wy .\}}\\ \cmidrule(r){2-3}
         &primitive  &\makecell[l]{[\textit{entity prim}] ns:people.person, m.0gwm\_wy  [\textit{relation prim}] ns:film.director.film}  \\ \cmidrule(r){2-3}
         &composition  &\makecell[l]{$N(x_1) \;\wedge\; R(x_1, x_2)$ } \\ 
        \bottomrule
    \end{tabular}}
    \captionof{table}{Examples from SCAN, COGS, GeoQuery, and CFQ illustrating primitive–composition decomposition. For each dataset, we show the original input–output pair and its decomposition into primitives and abstract composition structures. This formulation separates lexical primitives from structural composition, enabling compositional-level reward. We abstract primitives into entities/actions and relations, and represent compositional structures using unified predicate templates $N(x)$, $R(x,y)$, and $V(x)$.}
    \label{tab:datasets}
\end{strip}


\section{Data}
\subsection{Data Statistic}
\label{app:statistics}
We use four compositional benchmarks in this paper, including SCAN, COGS, GeoQuery, and CFQ. The data statistic is collected in Table \ref{tab:statistics}.
\begin{table}[h]
\centering
\begin{tabular}{l l r r r}
\toprule
Dataset & Setting & \# train & \# dev. & \# test \\
\midrule
\multirow{2}{*}{SCAN}
 & turnleft & 21890 & - & 1208 \\
 & length & 16990 & - & 3920 \\ \midrule
COGS & - & 24155 & 3000 & 3000 \\\midrule
\multirow{2}{*}{GeoQuery}
 & template & 544 & 60 & 276 \\
 & length & 540 & 60 & 280 \\\midrule
\multirow{3}{*}{CFQ}
 & MCD1 & 95743 & 11968 & 11968 \\
 & MCD2 & 95743 & 11968 & 11968 \\
 & MCD3 & 95743 & 11968 & 11968 \\
\bottomrule
\end{tabular}
\caption{Statistics of datasets and splits used in our experiments.}
\label{tab:statistics}
\end{table}

\subsection{Data Example for Composition Reward}
\label{app:compos_example}
Table \ref{tab:datasets} lists examples from four compositional generalization benchmarks, together with the primitive and compositional abstractions used to compute the compositional reward.

\end{document}